\documentclass{article}

\usepackage[preprint]{neurips_2026}

\usepackage{neurips_2026}

\usepackage[utf8]{inputenc} 
\usepackage[T1]{fontenc}    
\usepackage{hyperref}       
\usepackage{url}            
\usepackage{booktabs}       
\usepackage{amsfonts}       
\usepackage{nicefrac}       
\usepackage{microtype}      
\usepackage{xcolor}         
\usepackage{booktabs}
\usepackage{multirow}
\usepackage{graphicx}
\usepackage{lipsum}
\usepackage{float}
\usepackage{xspace}
\usepackage{wrapfig}
\usepackage[table]{xcolor}
\usepackage{algorithm}
\usepackage{algpseudocode}
\usepackage{amsmath}
\usepackage{graphicx}
\usepackage{subcaption}
\usepackage{enumitem}

\newcommand{\nohistory}{\textit{{no-prior}}\xspace}
\newcommand{\whistory}{\textit{{raw-interaction}}\xspace}
\newcommand{\proposed}{\textsc{{Polar}}\xspace}

\newcommand{\smallsection}[1]{{\vspace{0.05in} \noindent \bf {#1.\hspace{5pt}}}}
\definecolor{DarkGreen}{RGB}{30,130,30}

\title{Personalizing Embodied Multimodal Large Language Model Agents over Long-term User Interactions}

\author{%
  Jeongeun Lee$^{1}$ \quad
  Chanyoung Park$^{2}$ \quad
  Dongha Lee$^{1}$\thanks{\; Corresponding author} \\
  $^{1}$Yonsei University \quad
  $^{2}$KAIST \\
  \texttt{\{ljeadec31, donalee\}@yonsei.ac.kr} \\
  \texttt{cy.park@kaist.ac.kr} 
}

\begin{document}

\maketitle

\begin{abstract}
  Multimodal large language model (MLLM)-based embodied agents have shown strong potential for solving complex tasks in physical environments.
However, personalized assistance requires more than following generic instructions or recognizing object at the category level.
In real-world scenarios, the intended target is often specified only implicitly through prior interactions, requiring agents to ground user-intended instances from personalized context accumulated over time.
In this work, we propose \proposed, a multimodal memory-augmented framework for personalized embodied agents over long-term user interactions.
\proposed organizes prior interactions into a  multimodal knowledge graph with semantic memory for personalized context, and episodic memory for past embodied experiences such as agent trajectories.
Therefore \proposed retrieves candidate object memories to ground the user-intended target instance and guide subsequent planning.
We evaluate \proposed across multiple MLLM backbones and diverse evaluation scenarios to study how prior interactions should be represented and used for long-term personalization. 
Results show that current inputs alone are insufficient for personalized instance grounding, while raw prior interactions are difficult to use directly due to their unstructured form. 
\proposed improves performance by converting prior interactions into task-relevant memory. 

\begin{figure}[h]
    \centering
    \vspace{0.2cm}
    \includegraphics[width=0.99\linewidth]{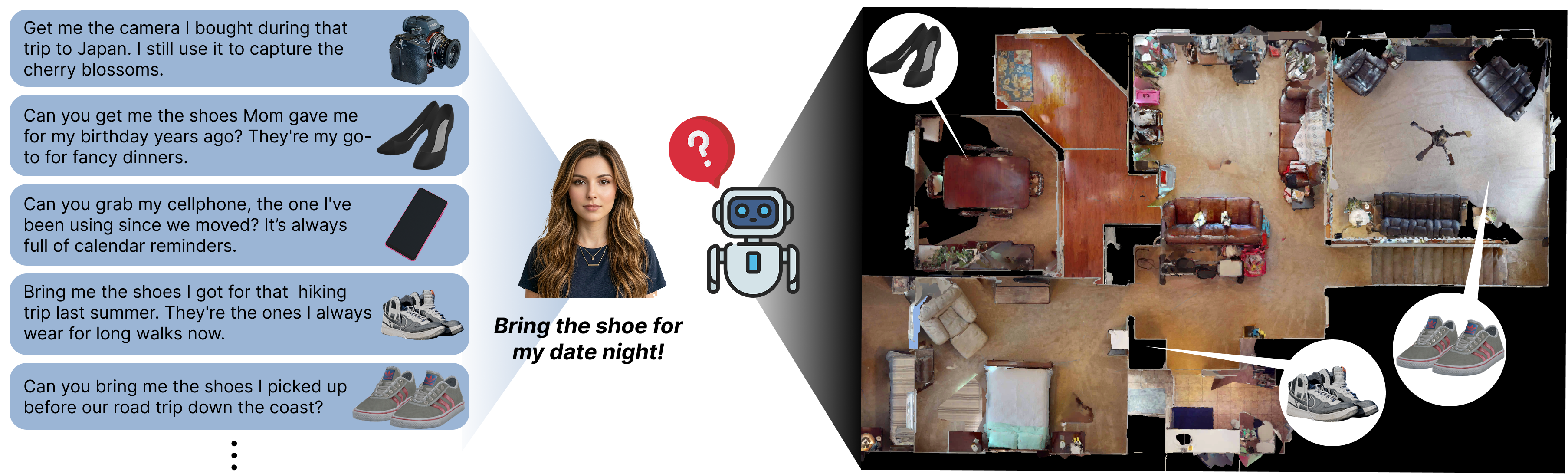}
    \caption{\textbf{Personalization over long-term user interactions.}
    In daily life, users often refer to objects through personal context accumulated over prior interactions rather than explicit target references.
    When multiple similar objects are present, conventional embodied agents may fail to determine which specific instance the user intends, as they focus on finding the category ``shoes'' rather than ``which'' shoes.
    This motivates personalized instance grounding, where embodied agents use long-term interaction history to resolve user intent.}
    \label{fig:intro}
\end{figure}
\end{abstract}

\section{Introduction}

With great advancements of multimodal large language models (MLLMs)~\cite{team2023gemini,hurst2024gpt,bai2025qwen3}, MLLM-based embodied agents have been increasingly studied for solving complex tasks while interacting with physical environments~\cite{mu2023embodiedgpt,driess2023palm, zitkovich2023rt,  szot2025multimodal,zhang2025vlabench}. 
By integrating visual perception and multimodal reasoning for vision-driven decision making,
these agents are expected to assist human in daily life, including locating~\cite{ayub2023personalized, ramrakhya2025grounding}, navigating to~\cite{zhao2025imaginenav, dai2024think, qiao2025open}, and delivering objects~\cite{tan2025language, korekata2026affordance} in response to user requests.

While prior works have largely focused on solving generic instructions, however, {category-level instance recognition} is not enough for personalized assistance.
As illustrated in Figure~\ref{fig:intro}, when a user asks for shoe under specific circumstances and multiple pairs of shoes exist in the house, the agent struggles to determine which pair to bring without personalized context.
Since such natural language-only instructions often fail to fully cover user request~\cite{lee2025bring}, some embodied agents learn to perform tasks on specific target instances with instance-level grounding~\cite{lei2024instance,barsellotti2024personalized, taioli2025collaborative}.
Still, these methods generally assume that the target reference is explicitly and directly given.
In real-world, however, such personalized context is more often {conveyed} implicitly through previous interactions, requiring the agent to interpret them to perform personalized instructions accordingly.

 \begin{figure}[b]
    \centering
    \vspace{-0.3cm}
    \includegraphics[width=0.99\linewidth]{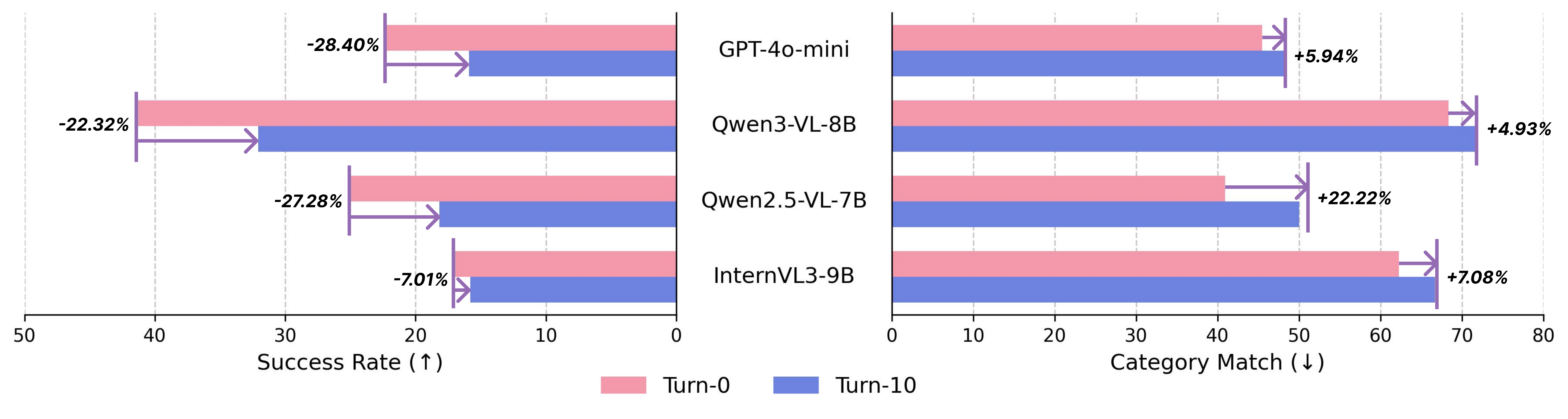}
    \vspace{-0.2cm}
    \caption{\textbf{Preliminary experiments on the PinNED dataset~\cite{barsellotti2024personalized}.} Success Rate measures correct target-instance navigation, while Category Match counts cases that reach the correct category object but the wrong instance. 0-turn and 10-turn denote the number of intervening interactions between the target reference and the final target instruction. Since the final instruction contains no reference, the agent must identify the target from prior interactions. We implement the agent in NavBench~\cite{qiao2026navbench}.}
    \label{fig:preliminary}
\end{figure}

Although leveraging long-term context is important for MLLM-based embodied agents, prior works have shown that MLLMs often degrade in long-context settings~\cite{kim2026mmpb,xue2025mmrc,bei2026mem}.
They often fail to reason effectively over accumulated multimodal context and instead rely on superficial cues~\cite{xue2025mmrc,bei2026mem, xu2026multihaystack}.
They also struggle to track how information evolves over time, making it difficult to capture updates or resolve conflicts across interactions~\cite{fu2026mmku}.
To further examine whether this challenge also arises in embodied multi-turn interactions, we provide a preliminary analysis in Figure~\ref{fig:preliminary}.
The results show that multiple MLLM-based embodied agents often fail to ground the intended target instance from prior interactions,
{
highlighting their difficulty in using accumulated user-specific information for personalized instance grounding.
}


For personalized embodied assistance, the agent should (1) manage information over long-term interactions to build personalized knowledge and (2) use that knowledge to interpret and adapt to individual user requests.
This requires more than simply remembering prior interactions, requiring the agent to organize them into task-relevant memory structures that can be selectively retrieved and applied to current instruction following and planning.
Therefore, in this work, we propose \proposed, a multimodal memory-augmented framework for \textbf{p}ers\textbf{o}na\textbf{l}ized MLLM-based embodied \textbf{a}gent for long-term use\textbf{r} interactions.
To this end,  \proposed leverages multimodal memory to accumulate personalized knowledge from prior interactions and retrieve relevant information for task execution.
While many existing memory methods primarily focus on compressing past interactions, the key idea of \proposed is to convert them into reusable personalized knowledge that can directly support future embodied tasks.
In particular, the memory organizes visual concepts together with personalized context (i.e., semantic memory) and agent trajectories (i.e., episodic memory)~\cite{tulving1972episodic}.
We organize these memories into a multimodal knowledge graph that connects semantic and episodic memories  across objects and interactions.


We study how \proposed structures prior interactions for long-term personalized embodied agents across multiple MLLM backbones and evaluation scenarios.
The results show that current observations alone are insufficient for personalized instance grounding. 
Raw prior interactions can contain useful user-specific information, but their unstructured form makes them {unstable} to use directly for the current task.
\proposed improves performance by converting prior interactions into task-relevant memory.
Further analyses show that semantic memory provides finer-grained retrieval of user-specific information, while episodic memory converts past trajectories into planning-relevant experience. 
These findings suggest that long-term personalization requires memory representations that support both personalized instance grounding and subsequent embodied planning.
\section{Related Works}

\smallsection{Vision-driven Embodied Agents}
Early works on embodied agents have primarily leveraged large language models (LLMs) with its high reasoning and planning capabilities~\cite{shridhar2020alfworld,huang2022language,huang2022inner,qian2024escapebench}.
They show that LLMs can decompose complex instructions~\cite{ahn2022can, cao2023robot,zhou2024navgpt}, reason over environment feedback~\cite{singh2022progprompt,  bhat2024grounding, kwon2024language}, and compose robot policies~\cite{kwon2024language, liang2023code, cheng2024empowering}.
However, these approaches typically relied on textual scene descriptions~\cite{zhou2024navgpt, zhang2024mapgpt, zhang2025vision}, external perception modules~\cite{huang2023voxposer}, or predefined skill libraries~\cite{ahn2022can, chu2024large}, limiting their ability to interpret subtle visual cue in raw observations.
To directly perceive fine-grained visual information, recent research has increasingly explored MLLM-based embodied agents, which integrate visual perception and language reasoning for vision-driven decision making~\cite{szot2025multimodal,qiao2026navbench,yang2025embodiedbench}.
These agents have demonstrated promising performance across a wide range of embodied tasks, including robot manipulation~\cite{driess2023palm, zhang2025vlabench}, household tasks~\cite{szot2025multimodal, ramrakhya2025grounding,xiao2024robi}, and embodied navigation~\cite{qiao2025open,qiao2026navbench,xu2025flame} in simulated and real-world environments~\cite{yang2025embodiedbench,zhang2025evaluating}.
Nevertheless,  as using MLLMs as step-by-step controllers over raw observations can be computationally inefficient for embodied control, particularly in long-horizon tasks~\cite{yue2024deer}.
Recent studies therefore leverage MLLMs as high-level planners, using them to predict abstract action plans, such as future waypoints~\cite{zhao2025imaginenav} or sequential action plans~\cite{yang2025embodiedbench}, rather than invoking them for per-step embodied decision making.

\smallsection{Personalized Embodied Agents}
Personalizing MLLMs aims to recognize user-specific visual entities beyond generic ones~\cite{kim2026mmpb,nguyen2024yo,alaluf2024myvlm}.
For example, while an off-the-shelf MLLM~\cite{team2023gemini,bai2025qwen3} may only recognize \textit{a dog}, a personalized MLLM~\cite{nguyen2024yo,oh2026repic} is expected to identify \textit{the user's dog}.
Beyond recognition, recent works have also explored retrieval-augmented personalization, where user-specific information is stored in an external database and dynamically retrieved to provide personalized context~\cite{hao2025rap,das2025training}.
The need for multimodal personalization has recently been extended to embodied agents, where agents are required to interpret user instructions by grounding user-specific objects in physical environments~\cite{dai2024think,lee2025bring,barsellotti2024personalized,taioli2025collaborative,ziliotto2025personal,wang2026user}.
This shifts the problem from generic object recognition to identifying which specific object instance is intended for the user.
However, existing approaches typically assume that such personalized references are explicitly available, whereas in realistic settings they are often implicit and must be inferred from prior interactions.

\smallsection{Memory-augmented Agents}
Memory modules support long-term interaction in (M)LLM agents beyond the limited context window~\cite{zhang2025survey}.
Early studies mainly developed memory systems for LLM agents~\cite{packer2023memgpt,chhikara2025mem0,kang2025memory,yu2026agentic}, where the agent actively manages memory operations such as storing and retrieving past information.
However, as these approaches are text-centric, recent work has therefore extended memory to incorporate multimodal observations~\cite{long2026seeing,liu2025memverse}.
Nevertheless, maintaining coherent long-term multimodal memory remains challenging, as agents still struggle to revise outdated memories, track evolving information, and resolve conflicts across interactions~\cite{xue2025mmrc,fu2026mmku,liu2025memverse}.


\section{Personalized Embodied Task over Long-term Interactions}

\subsection{Task Formulation}
We formulate our task as a partially observable Markov decision process (POMDP), defined by
$$
\mathcal{P} = (\mathcal{S}, \mathcal{A}, \Omega, \mathcal{T}, \mathcal{O}, \mathcal{I}, \mathcal{R}),
$$
where $\mathcal{S}$ denotes the state space, $\mathcal{A}$ denotes the action space, $\Omega$ denotes the observation space, $\mathcal{T}$ denotes the transition function, $\mathcal{O}$ denotes the observation function, $\mathcal{I}$ denotes the task instruction, and $\mathcal{R}$ denotes the reward function.

For each episode $k$, the agent is given a task instruction $\mathcal{I}_k$ and previous episodes $\mathcal{E}_{<k} = \{e_1, \dots, e_{k-1}\}$ for long-term memory retrieval, where each episode $e_i$ consists of a user instruction and the corresponding interaction trajectory.
At each time step $t$, the agent receives a partial visual observation $v_t \in \Omega$. Given the interaction history $\tau_t = (v_1, a_1, \dots, v_{t-1}, a_{t-1}, v_t)$, the MLLM policy $\pi$ predicts the next action conditioned on the current instruction and the retrieved memory:
$$
a_t = \pi(\tau_t, \mathcal{I}_k, \mathcal{E}_{<k}).
$$

In this work, we focus on navigation as the target embodied task.
This task requires the agent to ground the intended target instance using prior interactions and move toward it in a physical environment.
It also avoids additional confounding factors from manipulation or tool-use skills, allowing to study personalized target interpretation and spatial decision making in a controlled setting.

\begin{figure}[t]
    \centering
    \includegraphics[width=0.99\linewidth]{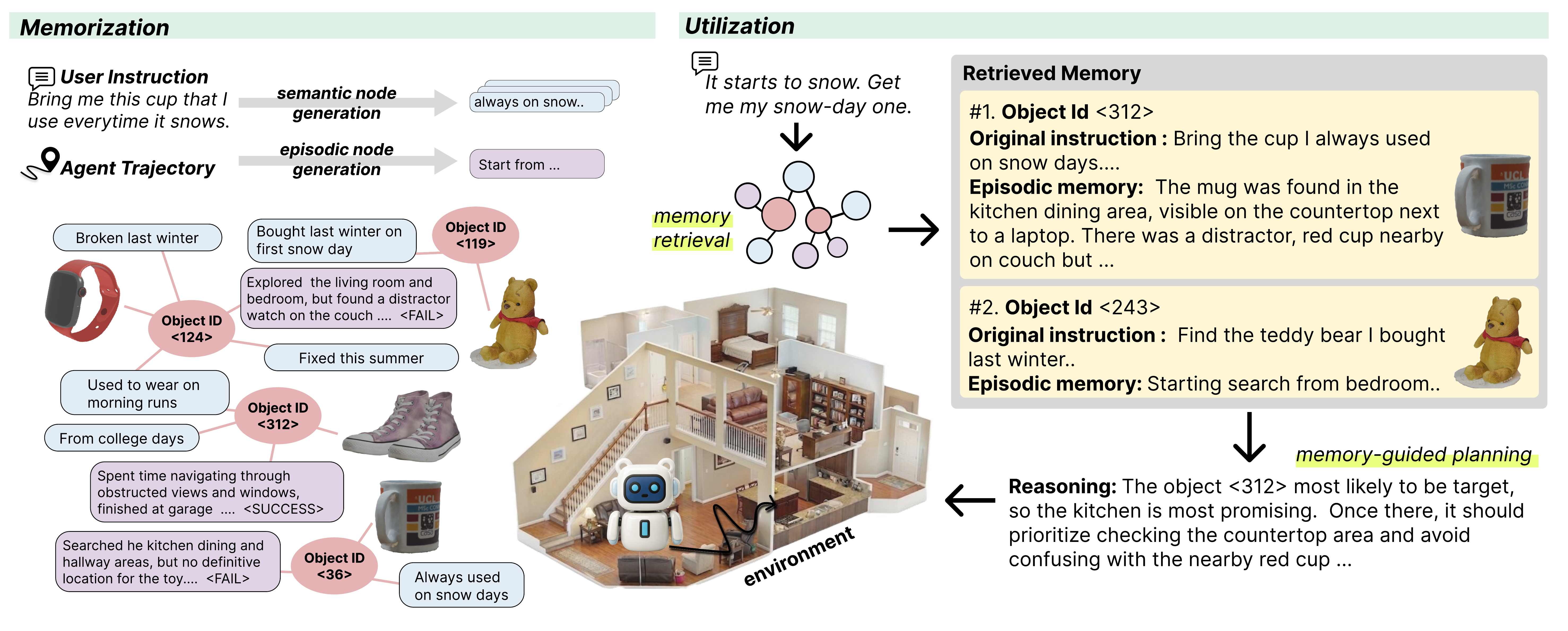}
    \caption{Overview of \proposed. In the \textbf{memorization} stage (left), \proposed builds an object-centric memory graph with semantic memory for personalized context and episodic memory for past trajectories. In the \textbf{utilization} stage (right), \proposed retrieves relevant memories for candidate objects to ground the target and guide subsequent planning. Timestamps are omitted for brevity.} 
    \label{fig:memory}
\end{figure}

\subsection{\proposed: Memory-augmented framework for Embodied Agents}
Figure~\ref{fig:memory} illustrates the \proposed framework, which consists of two stages: memorization and utilization.
Given prior episodes $E_{<k}$, the memorization stage converts previous interactions into the memory module, as $E_{<k}\rightarrow\mathcal{M}$.
Rather than storing interactions as raw logs or simple summaries, \proposed organizes them into personalized task memory associated with objects.
During utilization, \proposed retrieves memories conditioned on the current instruction $\mathcal{I}_k$,  as $\mathcal{R}_k=\rho(I_k,\mathcal{M})$.
The retrieved memory set $\mathcal{R}_k$ is then used by the embodied policy to predict the next action, $a_t=\pi(\tau_t,I_k,\mathcal{R}_k)$.
In this way, \proposed connects prior interactions to current task execution through task-relevant memory.



\subsubsection{Memorization}
We represent memory as a multimodal knowledge graph,
$\mathcal{M} = (\mathcal{V}, \mathcal{E}),$
where $\mathcal{V}$ denotes the set of memory nodes and $\mathcal{E}$ denotes the set of edges between them.
The node set $\mathcal{V}$ contains three types of nodes: object nodes, semantic memory nodes, and episodic memory nodes. 
Each object node corresponds to a unique object identity, represented by an object ID.
For each episode, \proposed generates two types of memory: semantic memory and episodic memory.

\textbf{Semantic memory} captures personalized knowledge associated with an object.
\proposed reformulates it into a set of concise semantic statements, where each statement preserves a single user-specific attribute or association about the target object.
This fine-grained formulation avoids entangling multiple pieces of personalized information in a single memory entry, which can weaken retrieval accuracy~\cite{chen2024dense}.
It also helps organize memory across objects and supports selective updates over time.
For object $o_i$, \proposed constructs
$
\{ s_{i,1}, s_{i,2}, \dots, s_{i,K_i} \},
$
where each $s_{i,k}$ is a short textual statement describing user-specific knowledge about $o_i$.

\textbf{Episodic memory} stores the agent trajectory together with the raw user instruction and, when available, the reference image.
As raw trajectories are often long and noisy, \proposed converts the full trajectory into a compact textual memory $p_i$.
Rather than preserving the entire episode in detail, $p_i$ captures planning-relevant experience, including whether the trajectory succeeded or failed, which rooms were unpromising, and which transitions or observations were informative.
This allows episodic memory to support future planning by reusing prior search experience~\cite{liu2026exploratory}.

After generating new memories, \proposed links them to an existing object node if the object ID matches or the reference image is visually similar to an existing one; otherwise, it creates a new object node.
Edges are timestamped to support temporal updates and recency-aware retrieval.
To avoid redundancy, if a newly generated semantic memory is sufficiently similar to an existing semantic memory node, \proposed links the corresponding object node to the existing node instead of creating a duplicate.
As a result, \proposed incrementally builds an object-centric structured memory that preserves personalized context over long-term interactions and supports effective retrieval for future tasks.


\subsubsection{Utilization}
Given an instruction $\mathcal{I}$, \proposed retrieves the top-$k$ semantic memory nodes from $\mathcal{M}$ using embedding similarity,
$
\mathrm{sim}(\mathcal{I}, s) = \phi(\mathcal{I}) \cdot \phi(s),
$
where $\phi(\cdot)$ is the BGE-M3 encoder~\cite{chen2024m3}.
We set $k=5$ to balance memory recall and retrieval noise, as larger values may introduce irrelevant context that confuses object grounding.
For each retrieved semantic node, \proposed follows its graph connection to the corresponding object node $z_j$, and gathers the associated instruction and episodic memory $p_j$.
\proposed then reasons over the retrieved candidate objects  with the current instruction, grounds the intended target, and leverages the corresponding episodic memory to guide planning.

\begin{figure}[h]
    \centering
    \vspace{0.1cm}
    \includegraphics[width=0.99\linewidth]{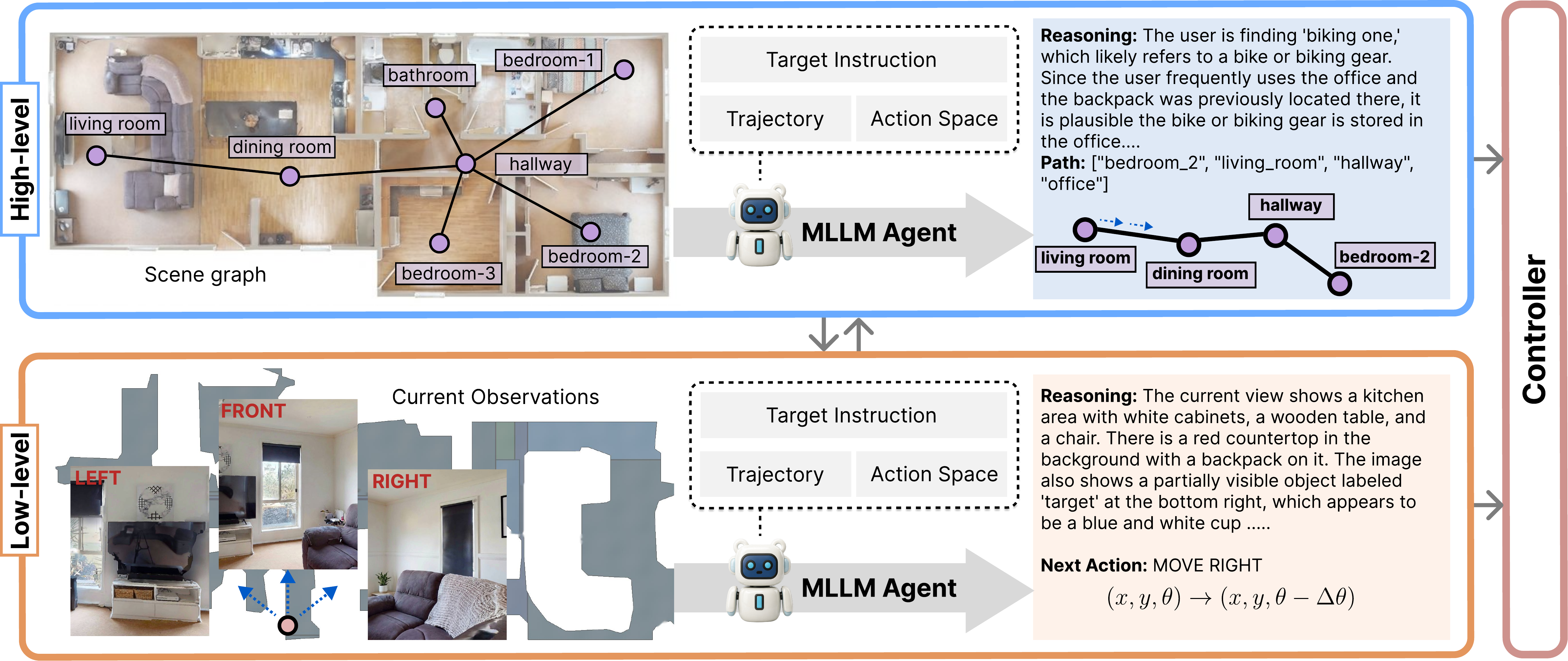}
    \caption{\textbf{Overview of the baseline MLLM embodied agent.} The agent follows a hierarchical planning framework with high-level planning (top), which predicts a coarse graph-based path toward a likely destination, and low-level planning (bottom), which selects executable actions from egocentric observations. The controller grounds these planned actions in the physical environment.}
    \label{fig:mllm_agent}
\end{figure}

\subsection{MLLM Embodied Agent}

\smallsection{Action Space}
Motivated by \cite{yang2025embodiedbench}, we model the action space hierarchically with two levels: high-level and low-level.
The low-level actions are directly executable actions, defined as
$$
\mathcal{A}_{\mathrm{low}} = \{\texttt{MOVE FORWARD}, \texttt{TURN LEFT}, \texttt{TURN RIGHT}, \texttt{STOP}\}.
$$
A high-level action specifies movement toward a selected waypoint or destination. This abstract navigation decision is further translated into a sequence of low-level actions,
$
[a_t, a_{t+1}, \dots, a_{t+n}],
$
which are executable in the environment.


\smallsection{High-level Planning}
Since searching for a target in a spacious environment cannot be efficiently solved from local observations alone~\cite{qiao2026navbench,yang2025embodiedbench}, the high-level planning module provides coarse guidance for long-horizon exploration.
Before task execution, the agent first traverses the environment to build a coarse spatial layout by identifying distinct regions and assigning region labels (e.g., living room)~\cite{zhang2024mapgpt,ziliotto2025personal,yin2024sg}.
It then converts this layout into a scene graph $\mathcal{G}$, where nodes correspond to rooms and edges represent adjacency between neighboring rooms.
Each node contains a waypoint that specifies a navigable target position within the corresponding room.
During exploration, given the scene graph and task context, the agent reasons over the global spatial structure to generate a promising movement directive toward the room most likely to contain the target.
The resulting waypoint sequence is then passed to the low-level module for execution.


\smallsection{Low-level Planning}
The low-level planning module directly interacts with raw visual observations to make short-horizon navigation decisions.
It takes multimodal inputs, including a natural language instruction, egocentric visual observations, and optionally a reference image of the target object.
At each step, the agent observes front, left, and right views from its current position, allowing it to compare candidate movement directions using local spatial evidence.
Based on these observations, it selects the next movement direction. When the target object is detected in any of the current views, the agent selects \texttt{STOP}.

\smallsection{Controller}
The controller executes primitive actions in the Habitat simulator~\cite{puig2023habitat3}. Specifically, \texttt{MOVE FORWARD} moves the agent by 1 meter, while \texttt{TURN LEFT} and \texttt{TURN RIGHT} rotate the agent by 30 degrees, grounding planned actions into physical movement. 

\section{Experiments}

\subsection{Experimental Settings}

\subsubsection{Dataset and Task Setup}

\smallsection{Dataset Construction}
We build our evaluation setup on top of PinNED~\cite{barsellotti2024personalized}, a personalized instance grounding dataset constructed in Habitat-Matterport3D (HM3D) scenes~\cite{ramakrishnan2021habitat} with photorealistic 3D objects.
We use 4 scenes with 1,817 episodes in total.
Following~\cite{kwon2026embodied}, we generate personalized context using GPT-5.\footnote{We perform a two-step filtering process: manual screening and then  MTurk-based validation of quality criteria, including naturalness, personalization clarity, ambiguity, and target consistency.} 
For each target object, we generate an explicit instruction containing the personalized context and an  evaluation instruction that requires the agent to infer the target from prior interactions. 
In the acquisition stage, the agent executes the explicit instruction to construct prior user--agent interactions. 
Although both instructions refer to the same target object, we vary object locations and agent initial positions across stages to prevent direct trajectory reuse.

\smallsection{Evaluation Scenarios}
We design three evaluation scenarios to assess different challenges of personalized interpretation under long-term interactions.

\begin{itemize}[leftmargin=1.2em,itemsep=2pt,topsep=2pt]
    \item \textbf{Compositional Scenario.}
    We consider two variants under the compositional scenario: \textit{single} and \textit{joint}. In the {single} setting, the target can be identified from a single prior episode. In the {joint} setting, the agent must combine information from multiple (2–3) prior interactions to determine which object is intended. This evaluates whether the agent can compose distributed personalized evidence across episodes to identify the correct target.

    \item {\textbf{Distractor Scenario.}}
    The environment contains multiple candidate objects from the same category as the target. This makes coarse category-level grounding insufficient, and requires the agent to distinguish the intended target from semantically or visually similar distractors.
    
    \item \textbf{Temporal Scenario.}
    The agent must capture for how personalized context changes over time. We consider two forms of change. First, the context associated with a specific object may evolve as the user’s routines or preferences change. Second, the object referred to by the same context may itself change over time. This scenario tests whether the agent can use the most recent valid interactions rather than rely on outdated associations.
    
\end{itemize}

\subsubsection{Evaluation Protocol}

\textbf{MLLM Backbones. }
We evaluate across multiple backbone MLLMs, including the open-source models Qwen3-VL-8B-Instruct and Qwen2.5-VL-8B-Instruct~\cite{bai2025qwen3}, as well as the proprietary models GPT-5, GPT-4o-mini~\cite{hurst2024gpt}, and Gemini-2.5-Flash~\cite{team2023gemini}.

\begin{figure}[t]
    \centering
    \includegraphics[width=0.99\linewidth]{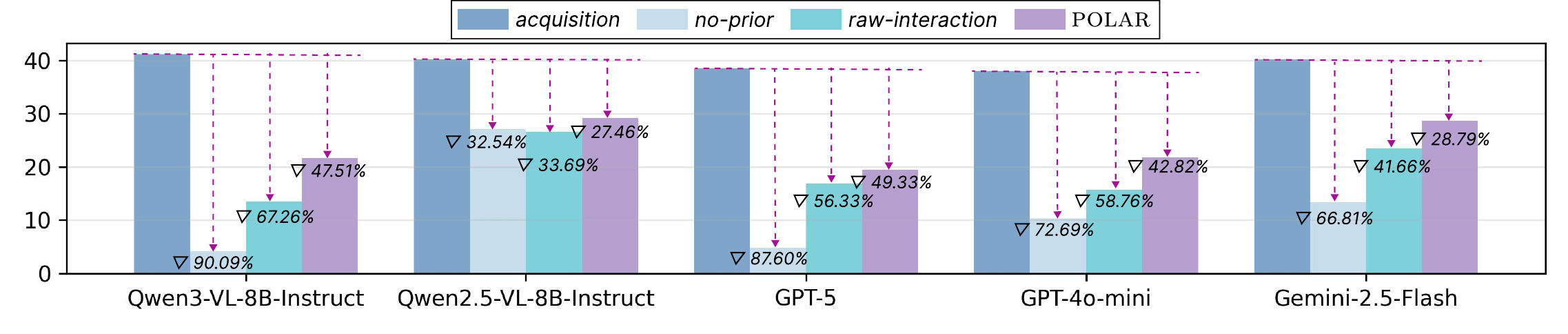}
    \vspace{-0.2cm}
    \caption{Performance gap between the acquisition and evaluation stages. In the acquisition stage, explicit personalized instructions with references are provided. \proposed yields the smallest gap when resolving underspecified instructions from prior interactions.}
    \label{fig:acq}
\end{figure}

\textbf{Evaluation Setup. }
Before evaluation, we first run the acquisition stage, where the embodied agent executes the explicit personalized instructions.
The resulting episodes, consisting of user instructions and agent trajectories, are then used as prior user--agent interactions across evaluation scenarios within each scene.
We compare the result from acquisition and evaluation stage in Figure~\ref{fig:acq}.

\textbf{Baselines. }
Based on these prior interactions, we compare \proposed with two baselines: (1) a {\textit{no-prior}} baseline, which uses only the current instruction and observations, and (2) a {\textit{raw-interaction}} baseline, where the agent is additionally given prior raw interactions. Because it is infeasible to include all prior interactions due to context limitations, we randomly sample 15 prior interactions, including the gold interaction, for the raw-interaction baseline.

\smallsection{Evaluation Metrics}
We use two widely-adopted metrics for embodied navigation~\cite{zhao2025imaginenav,qiao2026navbench,yang2025embodiedbench} : success rate (SR) and success weighted by path length (SPL)~\cite{anderson2018evaluation}. SR measures whether the agent successfully reaches the target object. An episode is considered successful if the distance between the agent’s final position and the goal object is less than two meters. SPL further accounts for navigation efficiency and is defined as
$
\mathrm{SPL} = \frac{1}{N} \sum_{i=1}^{N} S_i \frac{l_i}{\max(p_i, l_i)},
$
where $S_i$ is a binary success indicator for episode $i$, $p_i$ is the agent’s path length, and $l_i$ is the shortest-path distance to the goal object. We set the maximum number of steps to 700 for evaluation.

\begin{table*}[t]
\centering
\caption{{Performance comparison across multiple backbone MLLMs and evaluation scenarios.}}
\label{tab:main_results}
\resizebox{0.98\linewidth}{!}{%
\begin{tabular}{l|l|cccccccc}
\toprule
\multirow{3}{*}{Model} & \multirow{3}{*}{Setting}
& \multicolumn{4}{c}{Compositional}
& \multicolumn{2}{c}{Distractor}
& \multicolumn{2}{c}{Temporal} \\
\cmidrule(lr){3-6} \cmidrule(lr){7-8} \cmidrule(lr){9-10}
& & \multicolumn{2}{c}{Single} & \multicolumn{2}{c}{Joint}
& \multirow{2}{*}{SR} & \multirow{2}{*}{SPL}
& \multirow{2}{*}{SR} & \multirow{2}{*}{SPL} \\
\cmidrule(lr){3-4} \cmidrule(lr){5-6}
& & SR & SPL & SR & SPL & & & & \\
\midrule

\rowcolor{gray!15}
\multicolumn{10}{l}{\textit{Open-source MLLMs}} \\

\multirow{3}{*}{Qwen3-VL-8B-Instruct}
& \nohistory         & 5.75  & 2.53  & 5.56  & 0.00  & 0.00  & 0.00  & 5.00  & 2.51  \\
& \whistory          & 18.05 & 14.55 & 18.05 & 14.55 & 0.00  & 0.00  & 17.81 & 15.51 \\
& \proposed          & 23.17 & \textbf{20.95} & 21.84 & 15.14 & 20.00 & 18.00  & 21.43 & 18.67 \\
\cmidrule(lr){1-10}

\multirow{3}{*}{Qwen2.5-VL-8B-Instruct}
& \nohistory         & 20.00 & 11.70 & 29.73 & 13.95 & 30.00 & \textbf{27.77} & 25.71 & \underline{24.19} \\
& \whistory          & 27.27 & 18.38 & \underline{30.59} & \textbf{28.12} & 20.00 & 9.06  & 25.64 & 21.71 \\
& \proposed          & \underline{28.75} & 15.93 & \textbf{31.51} & \underline{18.44} & \textbf{60.00} & \underline{26.00} & \underline{27.08} & \textbf{24.34} \\

\midrule
\rowcolor{gray!15}
\multicolumn{10}{l}{\textit{Proprietary MLLMs}} \\

\multirow{3}{*}{GPT-5}
& \nohistory         & 8.33  & 6.09  & 6.25  & 4.98  & 25.00 & 14.73 & 4.55  & 4.55  \\
& \whistory          & 24.39 & 11.76 & 18.42 & 0.00  & 25.00 & 11.20 & 7.69  & 3.29  \\
& \proposed          & 24.49 & 14.70 & 20.59 & 9.29  & \underline{37.50} & 14.73 & 13.51 & 11.77 \\
\cmidrule(lr){1-10}

\multirow{3}{*}{GPT-4o-mini}
& \nohistory         & 6.82  & 0.00  & 7.50  & 0.00  & 20.00 & 2.24  & 7.14  & 1.62 \\
& \whistory          & 13.50 & 4.65  & 17.39 & 14.66 & 20.00 & 2.24  & 11.76 & 2.97 \\
& \proposed          & 25.30 & \underline{20.82} & 20.00 & 18.10 & 30.00 & 3.36  & 11.54 & 2.92 \\
\cmidrule(lr){1-10}

\multirow{3}{*}{Gemini}
& \nohistory         & 10.00 & 8.30  & 16.67 & 10.78 & 10.00 & 7.70  & 23.33 & 19.93 \\
& \whistory          & 24.14 & 15.96 & 22.73 & 14.21 & 25.00 & 3.24  & 23.34 & 20.21 \\
& \proposed          & \textbf{30.56} & 19.81 & 26.67 & 9.56  & 25.00 & 12.95 & \textbf{28.12} & 12.15 \\

\bottomrule
\end{tabular}%
}
\end{table*}

\subsection{Comparison with Baselines for Long-Term Personalization}

We organize our evaluation around the following research questions:
\begin{itemize}[leftmargin=*, itemsep=0pt, topsep=0pt, parsep=0pt]
    \item \textbf{RQ1:} Do the current instruction and observations provide enough information for personalized instance grounding?
    \item \textbf{RQ2:} Are raw prior interactions sufficient for long-term personalized task execution?
    \item \textbf{RQ3:} How does the memory design of \proposed affect target grounding and planning?
\end{itemize}

\smallsection{Main Results}
Table~\ref{tab:main_results} presents the main evaluation results across multiple backbone MLLMs and evaluation scenarios.
The \nohistory baseline often performs worse than both \proposed and the \whistory baseline across all scenarios.
This suggests that the current observations alone often lack the information needed to interpret the target instruction, highlighting the importance of prior interactions for personalized task execution.
The \whistory baseline sometimes improves over \nohistory by leveraging user-specific information, but the gains are inconsistent.
This suggests that unstructured prior interactions can introduce distracting context, making them difficult to use reliably for the current task.
The advantage of \proposed is especially clear in the joint and temporal scenarios, showing that the object-centric graph helps connect information distributed across multiple prior interactions.
{
Furthermore, we analyze whether \proposed improves instance-level grounding beyond category-level navigation.
Following~\cite{barsellotti2024personalized}, we report Category Match (CM), which measures within-category false positives where the agent reaches the correct category but the wrong target instance.}
Lower CM indicates fewer same-category target errors, and \proposed consistently yields lower CM than the \whistory baseline, showing that it better resolves the user-intended instance among same-category candidates.

\begin{figure}[h]
    \centering

    \begin{minipage}[t]{0.32\textwidth}
        \centering
        \includegraphics[width=\linewidth]{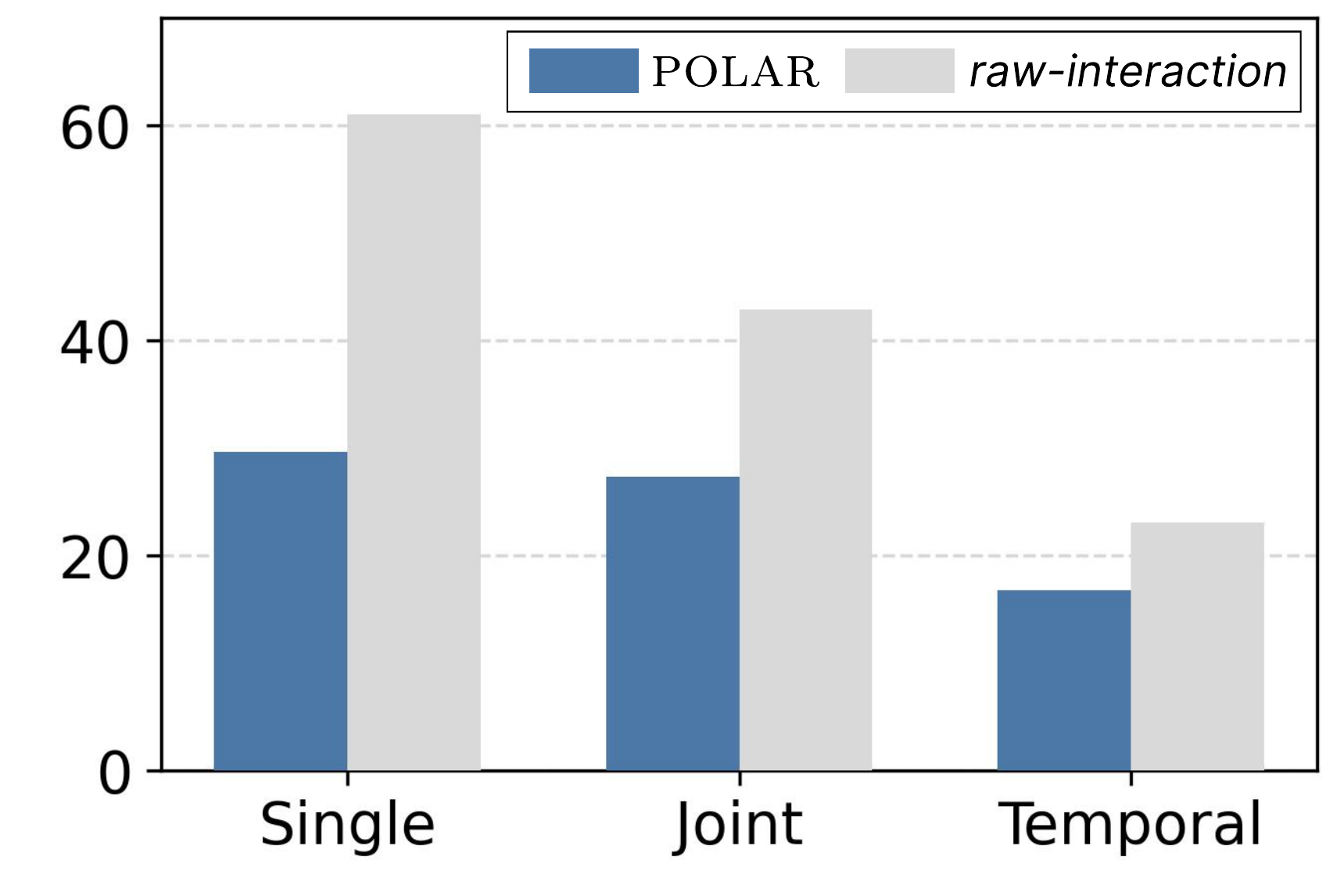}
        \vspace{-0.6cm}
        \captionof{figure}{Within-category false positive (i.e., CM).}
        \label{fig:category_match}
    \end{minipage}\hfill
    \begin{minipage}[t]{0.32\textwidth}
        \centering
        \includegraphics[width=\linewidth]{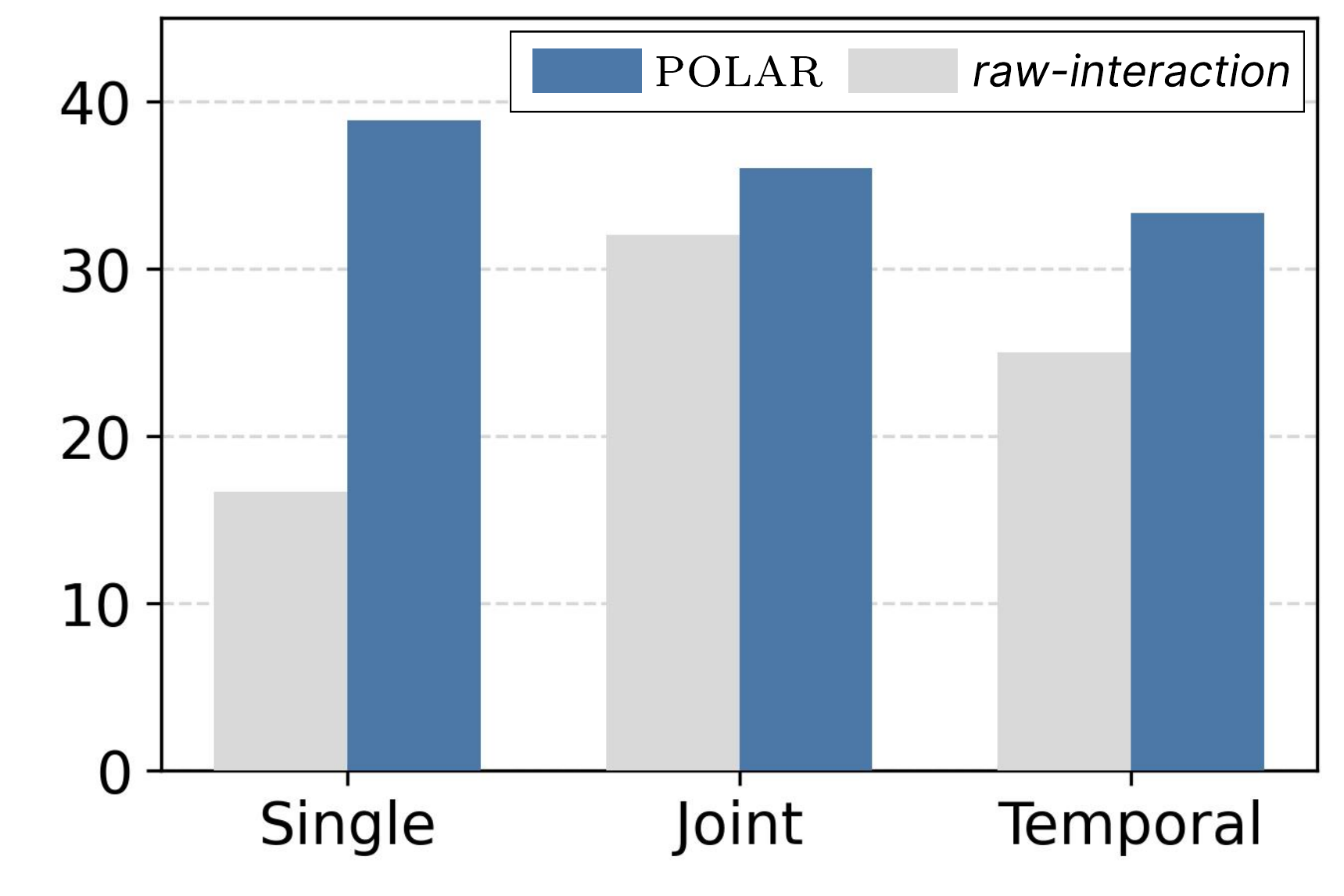}
        \vspace{-0.6cm}
        \captionof{figure}{Task success under a controlled retrieval setting.}
        \label{fig:sr_after_hit}
    \end{minipage}\hfill
    \begin{minipage}[t]{0.32\textwidth}
        \centering
        \includegraphics[width=\linewidth]{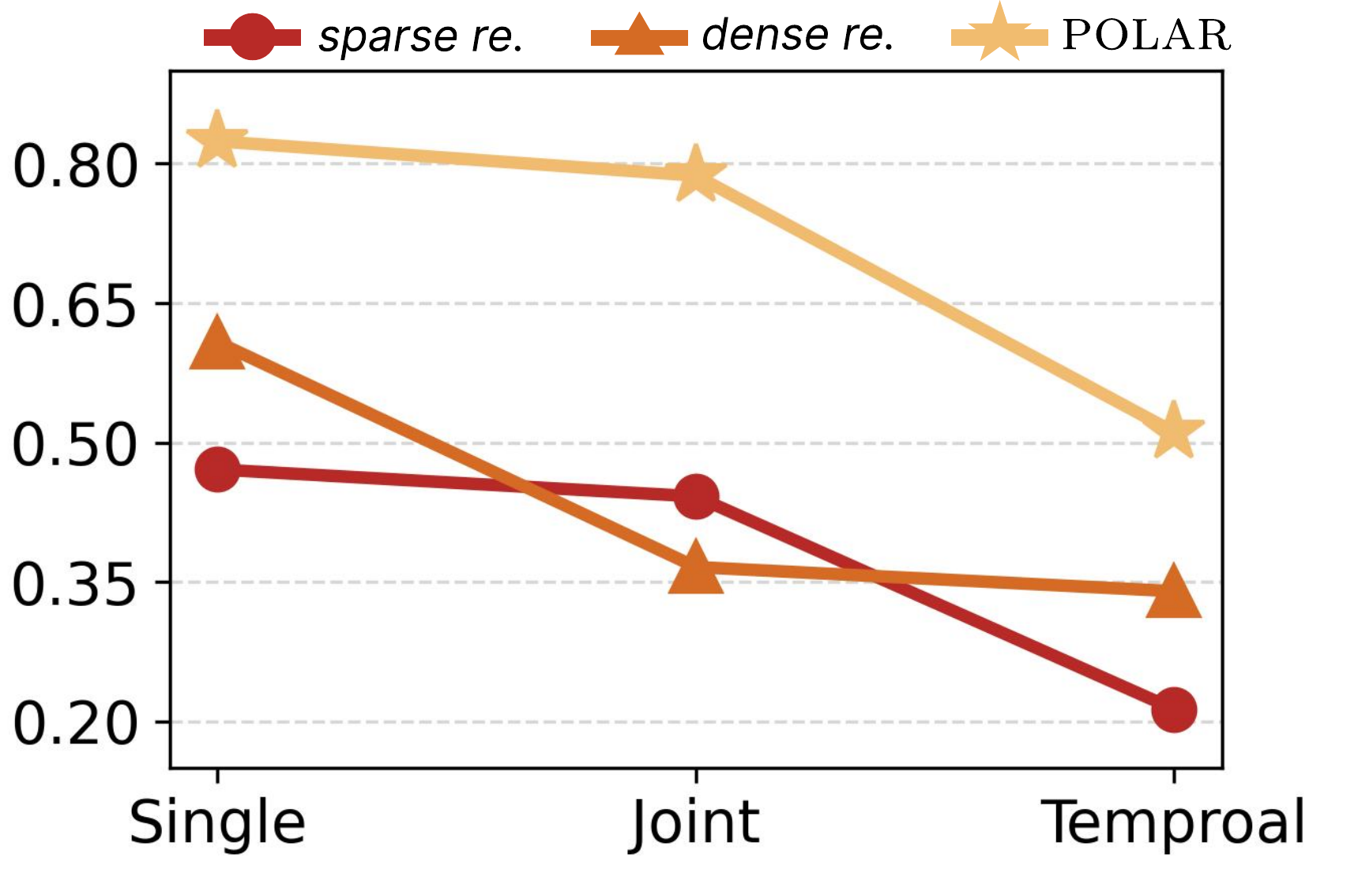}
        \vspace{-0.6cm}
        \captionof{figure}{Retrieval recall for task-relevant information.}
        \label{fig:retrieval_recall}
    \end{minipage}


\vspace{-0.5cm}
\end{figure}

\smallsection{Effect of Retrieved Memory Utilization} 
To further examine the effectiveness of the utilization process, Figure~\ref{fig:sr_after_hit} compares the success rate (SR) of \proposed and the raw-interaction baseline under a controlled retrieval setting, using cases where \proposed retrieves the target instance and the raw-interaction baseline includes the gold interaction.
\proposed achieves higher SR across all scenarios, showing that successful task execution requires more than retrieving the correct memory. The agent must also reason over the retrieved candidates to identify the intended target and apply the selected information to the current task.



\subsection{Analysis of Memory Design in \proposed}

\smallsection{Effect of Semantic Memory on Retrieval}
Figure~\ref{fig:retrieval_recall} analyzes how semantic memory affects retrieval of task-relevant personalized information from prior interactions.
We compare \proposed with sparse retrieval (BM25~\cite{robertson2009probabilistic}) and dense retrieval (BGE~\cite{chen2024m3}), both of which retrieve directly from raw prior interactions by matching them to the target instruction.
Yet \proposed retrieves semantic memory nodes, where prior interactions are reformulated into concise, object-specific statements. 
The higher recall of \proposed shows that raw interaction retrieval is too coarse-grained to isolate request-relevant information, while semantic memory provides a finer, representation.

\smallsection{Effect of Episodic Memory on Planning}
To examine the effect of episodic memory representation, we compare several memory representations under the same retrieval pipeline used in \proposed.
Recent memory-augmented agents often either preserve raw trajectories~\cite{feng2026m2a} or store
\begin{wraptable}{r}{0.35\textwidth}
    \centering
    \vspace{-0.4cm}
    \caption{Ablation on trajectory representations.}
    \label{tab:traj_ablation}
    \resizebox{\linewidth}{!}{%
    \begin{tabular}{lc}
        \toprule
        Setting & SR \\
        \midrule
        Instruction only & 31.1 \\
        + Raw trajectory & 31.9 \\
        + Summary & 34.5 \\
        \rowcolor{gray!15}
        + Episodic memory (Ours) & 36.2 \\
        \bottomrule
    \end{tabular}%
    }
    \vspace{-0.3cm}
\end{wraptable}
compact summaries of past interactions~\cite{liu2025memverse}.
Based on this design space, Table~\ref{tab:traj_ablation} evaluates
four variants: retrieved instructions only, raw trajectories, trajectory summaries, and the episodic memory used in \proposed.
The comparison shows that retaining more history alone does not necessarily improve performance.
As raw trajectories are long and include many task-irrelevant details, they can be difficult to exploit directly, whereas episodic memory extracts planning-relevant experience from past interactions,  enabling the agent to retrieve more actionable cues for the current task.


\begin{figure}[t]
    \centering
    \includegraphics[width=0.99\linewidth]{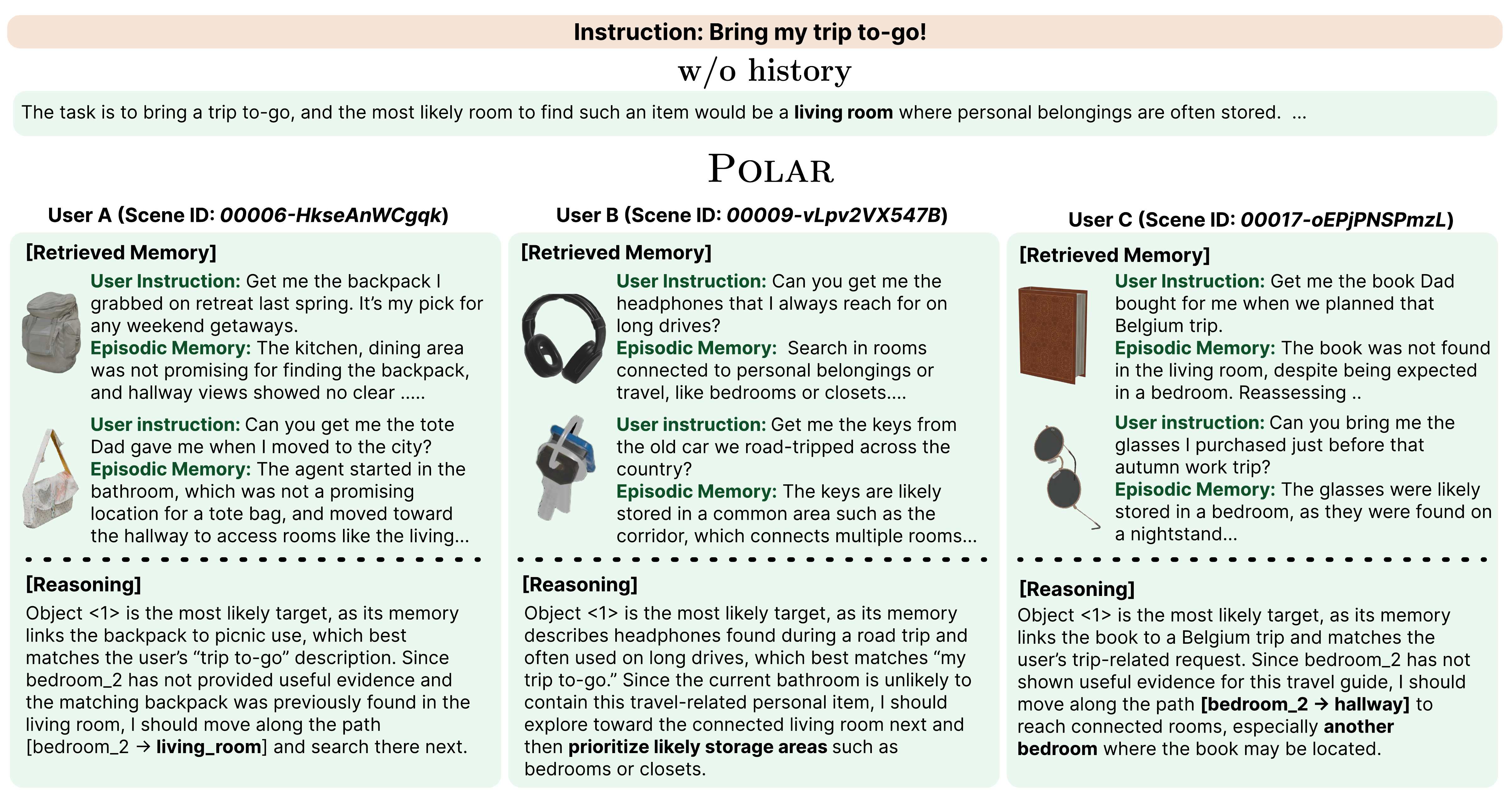}
    \caption{{\textbf{Case study of personalized embodied assistance.}}
    Given the same instruction, ``trip to-go,'' different users intend different target objects, such as a backpack, headphones, or a book, depending on their prior interactions. Without prior interactions, the agent relies on commonsense scene priors and gives the same generic answer, such as searching the living room. In contrast, \proposed retrieves user-specific memories to infer the intended target and to guide the planning.}
    \label{fig:case_study}
    \vspace{-0.4cm}
\end{figure}

\subsection{Case Study}
Figure~\ref{fig:case_study} provides a qualitative example of how the same instruction can lead to different behaviors depending on personalized memory.
Without prior interactions, the agent can only rely on generic commonsense and scene priors, which leads to similar interpretations across users.
On the other hand, \proposed uses retrieved semantic and episodic memories to condition both target grounding and subsequent planning on each user’s prior interactions.
This demonstrates that personalization affects not only target grounding, but also embodied search strategy.

\section{Conclusion}
In this work, we study long-term personalization for embodied agents, since real-world personalized assistance often requires agents to interpret user-specific context implicitly accumulated through prior interactions, rather than relying on explicit target references.
We propose \proposed, a memory-augmented framework that organizes personalized context and embodied experiences into multimodal memory.
Experiments across multiple MLLM backbones and scenarios show that \proposed outperforms no-prior and raw-interaction baselines.
These results highlight that effective long-term personalization requires not merely longer context, but structured memory that supports target interpretation and embodied planning.





\bibliographystyle{unsrt}
\bibliography{main}

\medskip






\end{document}